\documentclass[hidelinks,conference]{IEEEtran}
\usepackage{amsmath,amsfonts}
\usepackage{algorithm}
\usepackage{array}
\usepackage[caption=false,font=normalsize,labelfont=sf,textfont=sf]{subfig}
\usepackage{textcomp}
\usepackage{stfloats}
\usepackage{url}
\usepackage{verbatim}
\usepackage{graphicx}
\usepackage{cite}
\usepackage{algpseudocode}
\usepackage{booktabs}
\begin{document}

\title{Exploiting Boosting in Hyperdimensional Computing for Enhanced Reliability in Healthcare}

\author{SungHeon Jeong$^1$, Hamza Errahmouni Barkam$^1$, Sanggeon Yun$^1$,\\ Yeseong Kim$^2$, Shaahin Angizi$^1$ and Mohsen Imani$^1{}^{\star}$  \\ $^1$University of California, Irvine, CA, USA\\ $^2$DGIST, Daegu 42988, South Korea\\
$^{\star}$Corresponding Author: m.imani@uci.edu \vspace{-3mm}
}

\maketitle

\begin{abstract}
    Hyperdimensional computing (HDC) enables efficient data encoding and processing in high-dimensional spaces, benefiting machine learning and data analysis. However, underutilization of these spaces can lead to overfitting and reduced model reliability, especially in data-limited systems—a critical issue in sectors like healthcare that demand robustness and consistent performance. We introduce \textbf{BoostHD}, an approach that applies boosting algorithms to partition the hyperdimensional space into subspaces, creating an ensemble of weak learners. By integrating boosting with HDC, BoostHD enhances performance and reliability beyond existing HDC methods. Our analysis highlights the importance of efficient utilization of hyperdimensional spaces for improved model performance. Experiments on healthcare datasets show that BoostHD outperforms state-of-the-art methods. On the WESAD dataset, it achieved an accuracy of $98.37\% \pm 0.32\%$, surpassing Random Forest, XGBoost, and OnlineHD. BoostHD also demonstrated superior inference efficiency and stability, maintaining high accuracy under data imbalance and noise. In person-specific evaluations, it achieved an average accuracy of $96.19\%$, outperforming other models. By addressing the limitations of both boosting and HDC, BoostHD expands the applicability of HDC in critical domains where reliability and precision are paramount.
\end{abstract}

\section{Introduction}
    Hyperdimensional computing (HDC) is a rapidly advancing field within artificial intelligence, offering numerous valuable qualities that apply to various areas ~\cite{hd_9, hd_1, hd_2, hd_3, hd_8, hd_15, hd_16}. HDC's ability to efficiently encode and process data in high-dimensional spaces has generated substantial interest, particularly in machine learning and data analysis~\cite{Intro1, yun2023hyperdimensional, yun2024spatial, hd_4, hd_5, hd_6, hd_7, hd_10, hd_17, hd_18}. As we contemplate the practical use of HDC-based models in real-world applications, we encounter crucial requirements beyond mere accuracy. These requirements encompass robustness, reliability, consistency, and resource efficiency~\cite{onlineHD}. Despite these requirements, analyses of HDC in high-dimensional spaces has not yet been fully explored which can lead to overfitting by selecting very high dimension above only performance \cite{overfitting}. Our research shows that the partitioning of high-dimensional spaces directly correlates with the utility of the space, fulfilling those requirements under certain condition. 
    
    Meeting the requirements of HDC is anticipated to significantly expand its applicability across various domains, particularly in fields where reliability and precision are paramount, such as healthcare \cite{Intro2, Intro3}. In addressing healthcare datasets, our evaluation focuses on critical issues such as robustness to noise, consistency in performance, and stability that are non-negotiable for the successful application of HDC in healthcare. 
    
    Ensemble methods have been recognized for their ability to prevent overfitting and achieve enhanced performance and stability in predictive modeling~\cite{schapire2013explaining}. Among these methods, boosting stands out for its capability to aggregate the predictions of multiple weak learners into a robust and accurate ensemble model. The strengths of boosting encompass significant accuracy improvements, resistance to overfitting, versatility across a wide range of machine learning tasks. However, boosting`s ability across sensitivity to noisy data, performance and stability is dependent on the ability of weak learner. The computational demands of training boosting ensembles can also present challenges in scenarios requiring real-time responses or operating under resource constraints. Furthermore, the overall performance of the ensemble model may be compromised if the performance of the weak learners is not assured, potentially leading to biases towards challenging examples in imbalanced datasets.

    HDC offers promising attributes to effectively address the limitations associated with traditional boosting techniques, as highlighted in prior studies ~\cite{Intro7, Thomas_2021}. However, a simplistic parallel ensemble of HDC models may inadvertently escalate the computational costs associated with training and may not guarantee robustness against noise for each weak learner. In our approach, we leverage the OnlineHD model ~\cite{onlineHD} as a foundation, proposing a novel partitioning strategy where the model's hyperdimensional space ($D$) is divided among $n$ weak learners, with each receiving a $D/n$ dimensional segment. This segmentation approach prompts us to designate these segmented models as weak learners. Subsequently, we analyze the utilization efficiency of the hyperdimensional space by each weak learner Figure \ref{fig:span} and theoretically demonstrate the inherent limitations posed by high-dimensional spaces Equations \ref{eq: marchenko_std}, \ref{eq: marchenko_mean}. Under conditions where the performance of weak learners is assured, BoostHD elevates the capabilities of OnlineHD, ensuring stability, and providing robustness against overfitting and noise.

\section{Related Work}
    \begin{figure*}[t!]
        \centering
        \includegraphics[width=1\textwidth]{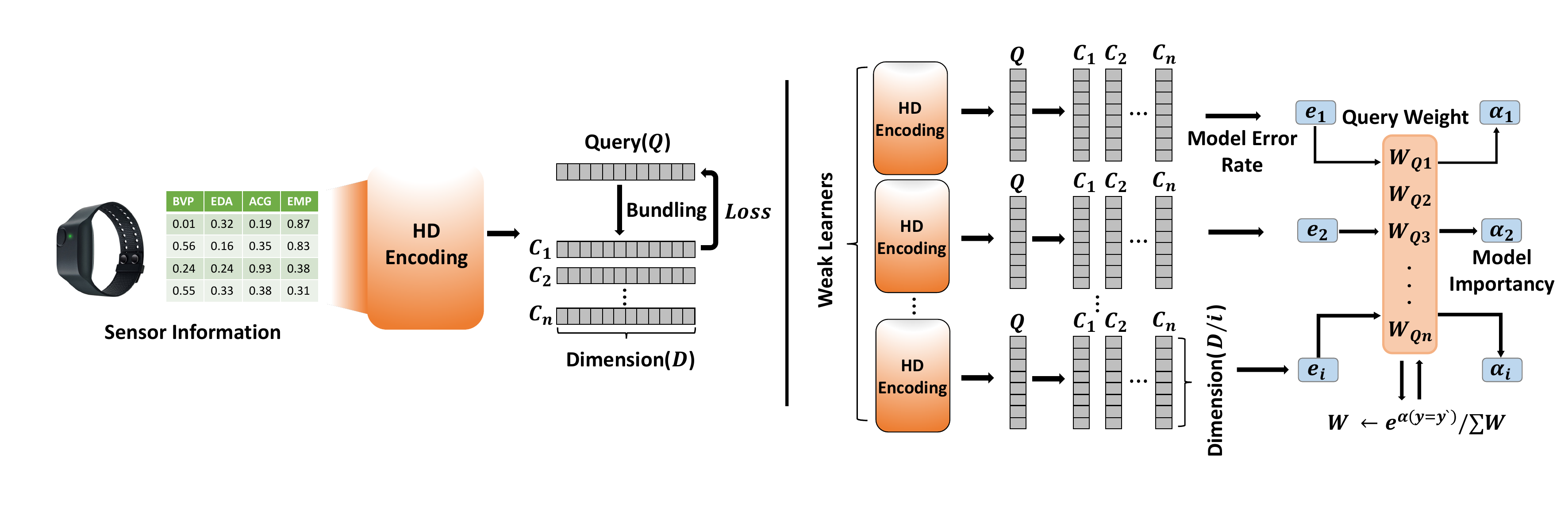}
        \caption{Illustration of the BoostHD framework applied to hyperdimensional computing (HDC). Sensor information is encoded into a high-dimensional vector space (Dimension $D$). The query vector $Q$ is bundled with multiple contextual vectors $C_1, C_2, \dots, C_n$, forming weak learners with segments of the high-dimensional space. Each weak learner receives a partitioned subspace ($D/n$) of the original hyperdimensional space, optimizing the use of the entire dimensional space and minimizing overfitting. Query weights $W_{Q1}, W_{Q2}, \dots, W_{Qn}$ and model importances are dynamically adjusted based on model error rates, with a boosting approach used to aggregate and adjust the ensemble performance, ensuring robustness and stability, particularly in noise-sensitive domains such as healthcare.}
        \label{fig:arch}
    \end{figure*}

    \subsection{Boosting methods on Machine Learning}
        To overcome the limitations of individual predictive models, ensemble methods such as boosting have been developed to improve performance by combining multiple weak learners, such as shallow decision trees or simple neural networks. Boosting aggregates the predictions of these learners, trained sequentially to correct each other's errors. A notable example is Adaptive Boosting (AdaBoost)\cite{schapire2013explaining}, which assigns equal initial weights to data points, adjusts weights to focus on misclassified points, and uses confidence-weighted voting for final predictions. Gradient Boosting Machine (GBM)\cite{friedman2001greedy}, in contrast, minimizes residual errors through gradient descent instead of reweighting data points. Extensions like LightGBM~\cite{ke2017lightgbm} and XGBoost~\cite{chen2015xgboost} adopt this residual minimization strategy, offering improved performance over AdaBoost.

    \subsection{Healthcare with wearable device}
        The WESAD dataset~\cite{wesad} is a benchmark for wearable stress and affect detection, comprising physiological and motion data from wrist- and chest-worn devices collected from 15 subjects in a controlled lab setting. It includes diverse sensor modalities such as blood volume pulse, electrocardiogram, electrodermal activity, respiration, temperature, and three-axis acceleration, alongside three affective states—neutral, stress, and amusement. While convolutional neural networks (CNNs) dominate as the state-of-the-art solution~\cite{cnn}, Hyperdimensional Computing (HDC) offers a resource-efficient alternative better suited for the constraints of wearable devices. HDC's lightweight, online learning approach effectively handles WESAD's multimodal data, making it a promising method for real-world applications in stress and affect detection. The combination of WESAD's rich modalities and HDC's computational efficiency underscores the dataset's potential as a testing ground for advanced wearable technologies.

    \subsection{Hyperdimensional Classification}
        HDC is inspired by neural processes in the brain, encoding data into a high-dimensional space. It learns universal patterns for each label by encoding data points as hypervectors \(\mathbf{H}\) through matrix multiplication with Gaussian distribution values and trigonometric activation functions (\textbf{Binding}), such as sine and cosine. These hypervectors are combined (\textbf{Bundling}) to create class hypervectors \(\mathbf{C}_l\), which represent each label in \(R^D\) space. During inference, a query hypervector \(\mathbf{H}\) is encoded and compared to class hypervectors using a similarity function \(\delta(\mathbf{H}, \mathbf{C}_l)\). This process is widely applicable, as demonstrated in ~\cite{hd_11, hd_12, hd_13, hd_14, hd_19, hd_20, hd_21, hd_22}.
        
        Model updates now leverage similarity assessments to iteratively refine predictions~\cite{onlineHD}, enhancing the model's ability to capture patterns and improve accuracy:  
        \begin{align}  
            \delta(V_1, V_2) = \frac{V_1^\dagger V_2}{\|V_1\|\|V_2\|}  
        \end{align}  
        
        Key operations in HDC include:  
        \begin{itemize}[leftmargin=*]  
            \item \textbf{Bundling}: Combines hypervectors into a single vector (\(\vec{R} = \vec{V}_1 + \vec{V}_2\)), enabling memorization in high-dimensional space.  
            \item \textbf{Binding}: Associates orthogonal hypervectors (\(\vec{R} = \vec{V}_1 * \vec{V}_2\)), creating a new hypervector orthogonal to inputs (\(\delta(\vec{R}, \vec{V}_1) \simeq 0\)).  
        \end{itemize} 

    \begin{algorithm}
        \caption{Pseudo code for BoostHD}\label{alg:pseudo_code_}
        
        \begin{flushleft}
            \quad \textbf{Input:}\\
                \quad \quad $X$: Set of data points\\
                \quad \quad $y$: Labels\\
                \quad \quad $x$: Test data point\\
            \quad \textbf{Parameters:}\\
                \quad \quad $d$: Dimensionality\\
                \quad \quad $n$: Number of learners\\
                \quad \quad $W_s$: Sample weight\\
            \quad \textbf{Output:}\\
                \quad \quad $f_{\theta_i}$: Trained learners parameterized by $\theta$\\
                \quad \quad $\alpha_i$: Weight of learner\\
                \quad \quad $\hat{y}$: Prediction
        \end{flushleft}
        
        \begin{algorithmic}[1]
        \Procedure{Training}{$X$, $y$}
            \State \textbf{Initialize learners} $f_{\theta_1}, f_{\theta_2}, \ldots, f_{\theta_n}$
            \For{$i \in \{1, 2, \dots, n\}$}
                \State Train $f_{\theta_i}$ with $X$ and $y$
                \State $\hat{y}_i = f_{\theta_i}(X)$
                \State $e_{\theta_i} \gets \text{Error rate of } f_{\theta_i}$
                \State $\alpha_i = W_s \cdot e_{\theta_i}$
                \State $W_s \gets e^{\alpha_i \cdot (y \neq \hat{y})} / \sum W_s$
            \EndFor
        \EndProcedure
        \end{algorithmic}
        
        \begin{algorithmic}[1]
        \Procedure{Inference}{$x$}
            \State $\hat{ys} = f_{\theta}(x)$
            \State $\hat{y} = \text{argmax} \left( \sum^n \hat{ys} \cdot \alpha \right)$
        \EndProcedure
        \end{algorithmic}
    \end{algorithm}

\section{Boosting Hyperdimensional Computing}
    In this section, we introduce the BoostHD framework, integrating boosting techniques into HDC with a focus on dimensionality ($D$). Instead of relying solely on a single robust learner with high $D$, this approach breaks down the dimension into numerous sub-dimensions, each represented by a weak learner. These weak learners are trained sequentially, with each one adaptively learning from and correcting the errors of its predecessor. This methodology effectively addresses the limitations of the strong learner, thereby enhancing its performance ceiling (as outlined in Algorithm \ref{alg:pseudo_code_}). A notable feature of this method is its sequential training process, while parallelization becomes feasible during the inference phase.

    Performance, assessed based on the parameters $D$ and the number of weak learners ($N_L$), shows a direct relationship, ensuring stable and improved performance with substantial values for both $D$ and $N_L$, as depicted in Figure \ref{fig:heatmap}. However, it's crucial to note that elevated values of $D$ and $N_L$ introduce increased computational costs, establishing an inherent tradeoff between computational cost and performance. To maintain the effectiveness of weak learners, preserving a baseline dimensionality is imperative.Failure to do so, as exemplified in the case where $N_L = 100$ and $D_{total}=1\text{K}$, can lead to a substantial degradation in performance, as demonstrated in Figure \ref{fig:heatmap}(b).
    \begin{equation}
        \begin{aligned}
            \mu_\lambda &= \int_{\lambda_{\text{min}}}^{\lambda_{\text{max}}} f(\lambda) \lambda \, d\lambda \sim \frac{1}{3\pi q} (\lambda_{\text{max}}-\lambda_{\text{min}})^{3/2}
        \end{aligned}
        \label{eq: marchenko_mean}
    \end{equation}
    \begin{equation}
        \begin{aligned}
            &\sigma^2_\lambda \sim \frac{1}{2\pi \sigma^2 q} \left( \frac{1}{2}(\lambda_{\text{max}}^2 - \lambda_{\text{min}}^2) - 2\mu(\lambda_{\text{max}}- \lambda_{\text{min}}) \right. \\
            &\qquad\qquad\qquad\qquad \left. + \mu^2(\ln |\lambda_{\text{max}}| - \ln|\lambda_{\text{min}}|) \right)
        \end{aligned}
        \label{eq: marchenko_std}
    \end{equation}

    In the context of HDC, partitioning dimensions can be viewed as a transformation in the geometric shape of the kernel ($k$). HDC frameworks often employ the Gaussian Kernel as their foundational element, represented as $k_{i, j} \sim \mathcal{N}(0, 1)$. When this kernel transforms a hyperdimensional space, the geometric characteristics of $k$ can be described by leveraging the theory of the Marchenko-Pastur distribution of random matrices (\cite{gotze2004rate}). Equations \ref{eq: marchenko_mean} and \ref{eq: marchenko_std} provide insights into this process. Here, $q$ is defined as the ratio of the number of columns ($N_c$) to the number of rows ($N_r$), $\sigma$ represents the standard deviation of the Gaussian distribution (i.e., 1), $\lambda$ denotes singular values, and $\lambda_{max}$ and $\lambda_{min}$ signify the upper and lower bounds of $\lambda$. The mean and variance of $\lambda$ are expressed in Equations \ref{eq: marchenko_mean} and \ref{eq: marchenko_std}, respectively. Considering that $N_c$ is determined by the dataset, and $N_r$ is a parameter represented by $D$, under the assumption of a fixed input shape, it becomes evident that $q$ exhibits an inverse relationship with $D$, while $\mu_\lambda$ demonstrates a direct proportionality with $D$. Conversely, Equation \ref{eq: marchenko_std} formulates $\sigma_\lambda^2$ as a function of three distinct terms: $T_1$ (Equation \ref{T1}), $T_2$ (Equation \ref{T2}), and $T_3$ (Equation \ref{T3}). Each term converges to a specific value and experiences minimal fluctuations after that. Consequently, $\mu_\lambda$ increases directly in proportion to $D$, while $\sigma_\lambda^2$ remains constant as $D$ increases. 

\begin{figure}[h]
        \centering
        \includegraphics[width=0.5\textwidth]{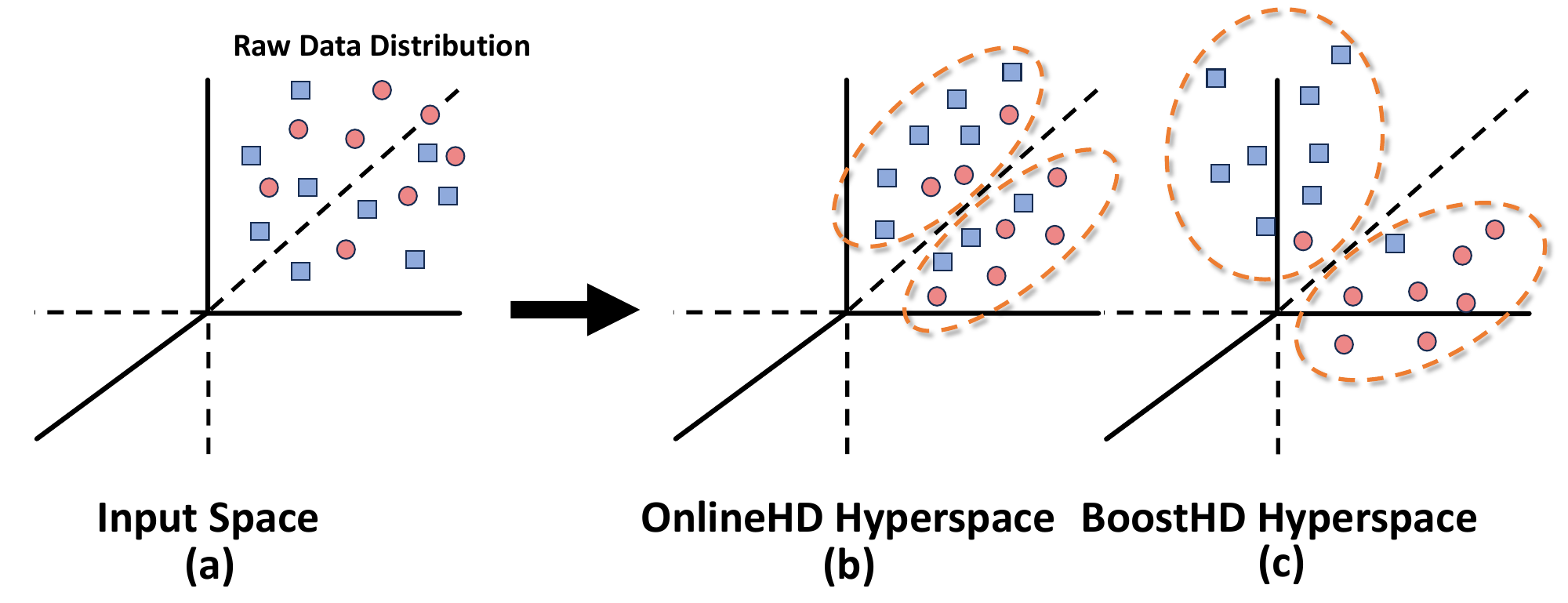}
        \caption{In the process of kernel transformation. Data is mapped into a hyperdimensional space. (a) illustrates the distribution of the raw data has a biased distribution. (b) represents a scenario where \(N_c=4000\), while (c) corresponds to \(N_c = 400\). From the perspective of span utilization, the mapping illustrated in (c) demonstrates superior efficiency compared to that in (b).}
        \label{fig:kernel_effect}
    \end{figure}

    This proportionality implies that irrespective of the values of $\mu_\lambda$ and $\sigma^2_{\lambda}$, $\lambda$ maintains a consistent absolute interval value due to the stability of $\sigma_\lambda^2$ (Equation \ref{T4}). As $D$ increases, the values of $\lambda$ escalate, yet the interval remains steady. This phenomenon results in the ratio of the minor axis ($A_S$) to the major axis ($A_L$) asymptotically approaching unity (i.e., 1), leading to a circular shape. This observation underscores the pivotal role of $D$ in shaping the kernel, wherein an elliptical kernel is transformed into a broadly distributed circular shape beyond the constrained space formed by input biases, as illustrated in Figure \ref{fig:kernel_effect}.

    \begin{gather}
        \lim_{q \to \infty} \frac{1}{q}(\lambda_{\text{max}}^2-\lambda_{\text{min}}^2) = \lim_{q \to \infty} \left((1+\sqrt{q})^4 - (q-\sqrt{q})^4 \right) = k \label{T1}\\
        \lim_{q \to \infty} \frac{1}{q} \left( -2\mu(\lambda_{\text{max}}-\lambda_{\text{min}}) \right) = 0  \label{T2}\ \\
        \lim_{q \to \infty} \frac{1}{q} \left( \mu^2(\ln |\lambda_{\text{max}}|-ln|\lambda_{\text{min}}|) \right) = 0 \ \label{T3} \\
        \lim_{x \to \infty}\sigma_\lambda^2 = k \label{T4}
    \end{gather}
    
    \begin{figure}[h]
        \centering
        \includegraphics[width=0.5\textwidth]{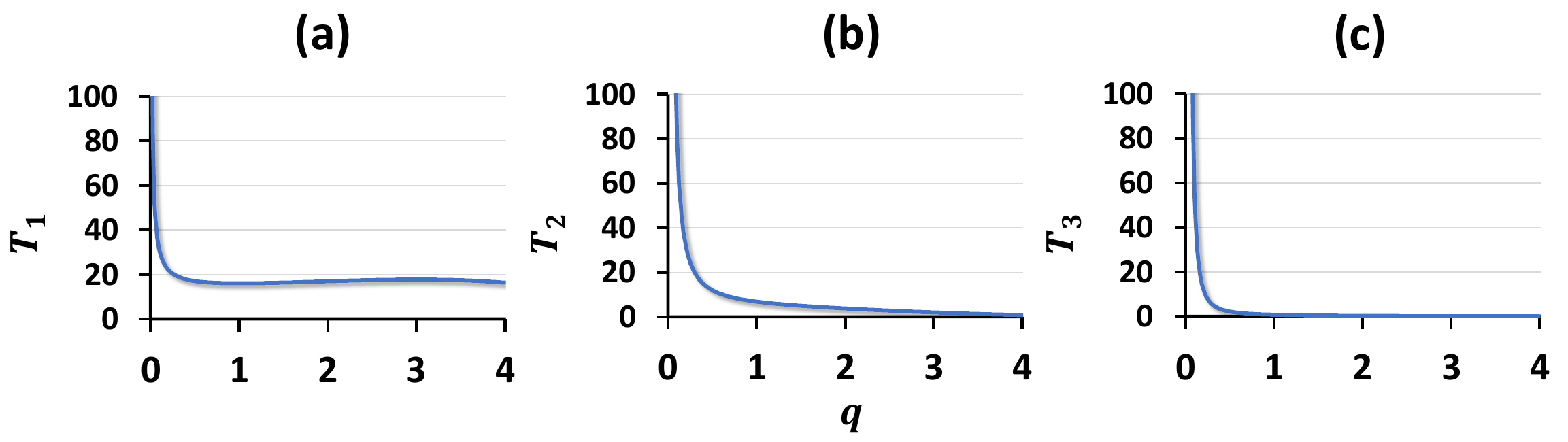}
        \caption{Extreme distribution of terms, Eq. \ref{T1}, \ref{T2}, \ref{T3} in $\sigma_{\lambda}^2$}
        \label{fig:heatmap}
    \end{figure}

    \begin{figure}[h]
        \centering
        \includegraphics[width=1\columnwidth]{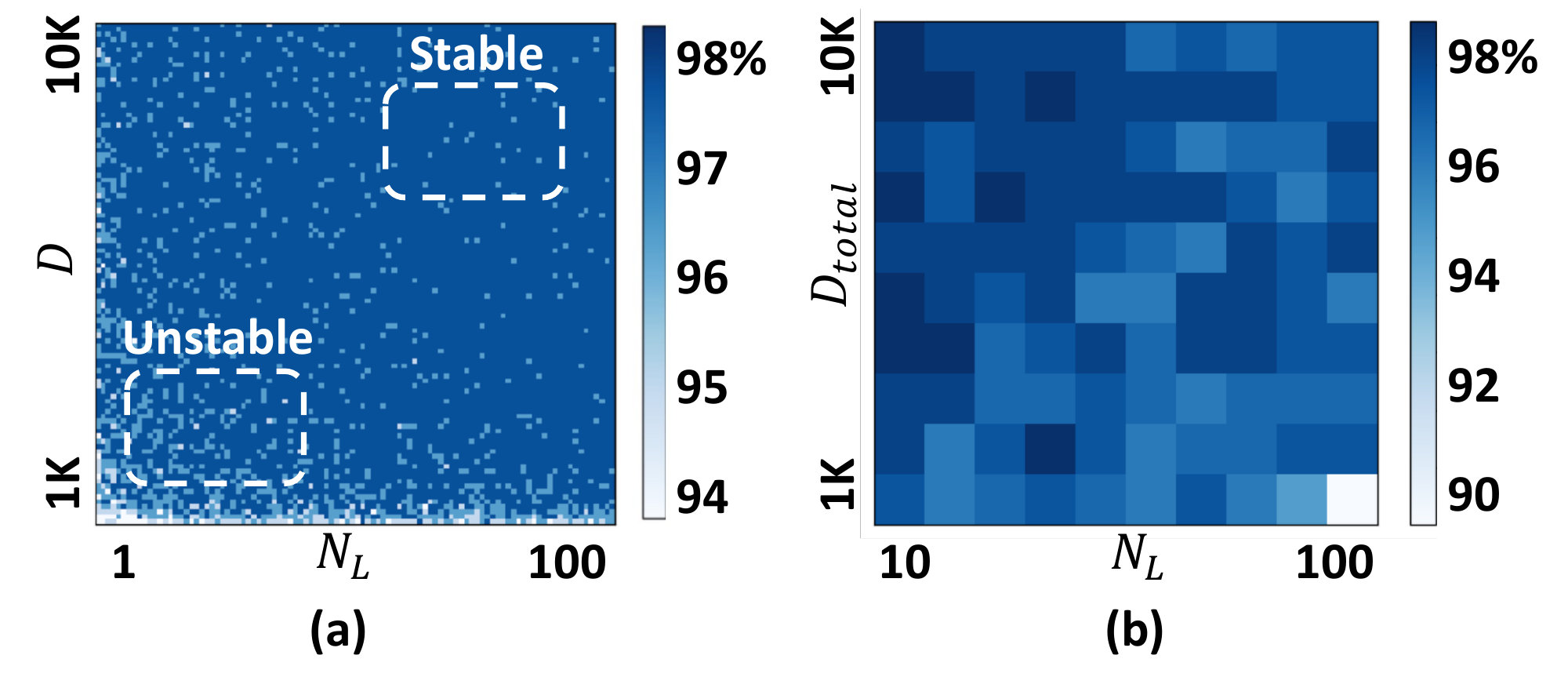}
        \caption{Accuracy heatmap based on $N_L$ and their respective $D$. In (a) and (b), $N_L$ takes values from 1 to 100 and 10 to 100 with each step 1, 10. For (a), the accuracy is presented for each specified dimension. For (b), the total dimension($D_{\text{total}}$) is divided among the $N_L$, where each learner possesses a dimension size of $D_{\text{total}} / N_L$.}
        \label{fig:heatmap}
    \end{figure}

\begin{figure}[h]
        \centering
        \includegraphics[width=0.5\textwidth]{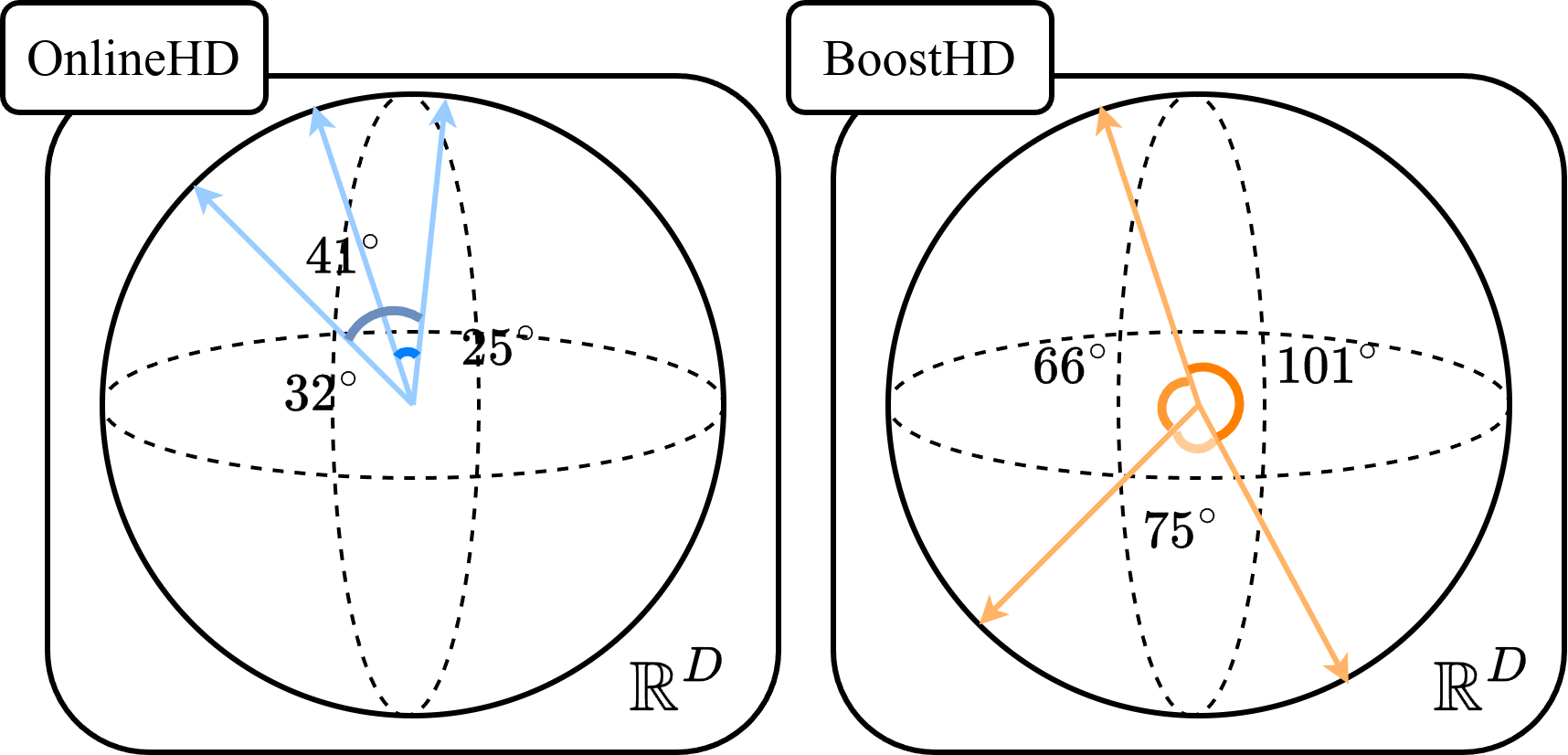}
        \caption{Example of BoostHD and OnlineHD $SP$. The orange one is BoostHD's class hypervectors, and the blue one is OnlineHD's class hypervector. BoostHD uses much more space at hyperdimensional space by composing high cosine similarity across class hypervectors.}
        \label{fig:span}
        
    \end{figure}

    Within the framework of HDC, the theoretical definition of utilizing the subspace formed by classifiers can be expressed as $\text{rank}(K) / D$, where $\text{rank}(K)$ represents the rank of the matrix formed by the classifier, and $D$ signifies the dimensionality of the space. However, in practical applications, the effective span of the subspace experiences attenuation due to various factors denoted by $\Pi$. These factors are the product sums of cosine similarity values between class hypervectors, which embody constraints or characteristics inherent to the system. Consequently, the actual Span Utilization ($SP$) is defined as the quotient of $\text{rank}(K)/D$ divided by the product of $\pi_1, \pi_2, ..., \pi_n$. This representation encapsulates the reduced space that is practically attainable, considering the influence of the aforementioned factors. In the realm of HDC, where cosine similarity serves as the metric of choice, maximizing the utilization of this subspace directly impacts the system's performance. With a focus on optimizing $SP$ (as depicted in Figure \ref{fig:span}), the BoostHD approach is superior to traditional HDC methodologies. By augmenting the subspace's utilization, BoostHD optimizes computational resources and significantly enhances the accuracy and efficiency of high-dimensional data processing. This observation underscores the practical relevance of subspaces in HDC and highlights the pivotal role played by BoostHD in maximizing their utility.

\section{Evaluation}
    In our experimental setup, we used a GeForce RTX 4070 GPU and a 12th generation Intel(R) Core(TM) i7-12700K CPU. Each experiment was conducted over 10 runs, employing datasets including WESAD~\cite{wesad}, the Nurse Stress Dataset~\cite{nursestress}, and the Stress-Predict Dataset~\cite{Stress-PredictDataset} to evaluate accuracy and time efficiency. For other aspects of the experiments, only the WESAD dataset was used. The WESAD dataset, a benchmark for stress detection research, features multimodal data from wearable devices like Empatica E4 and RespiBAN collected from 16 subjects. It includes physiological and motion indicators such as EDA, ECG, EMG, and BVP. The Nurse Stress Dataset~\cite{nursestress} and Stress-Predict Dataset~\cite{Stress-PredictDataset}, collected via the E4 sensor, also include EDA, ECG, EMG, and BVP data from 37 and 15 subjects, respectively. Stress level classification was reduced to three labels: good, common, and stress. Test data was organized by subject units, and all results were based on the test dataset.

    Datasets were preprocessed using a moving average filter with a window size of 30, extracting statistical features such as minimum, maximum, mean, and standard deviation. To address varying ranges, normalization was applied to ensure consistent scaling. The experimental models included AdaBoost (learning rate = 1.0, 10 estimators), Random Forest (bootstrap enabled, 10 estimators), XGBoost (10 estimators), SVM (linear kernel), and a DNN comprising convolutional layers with a learning rate of 0.001, four linear layers [2048, 1024, 512, classes], ReLU activation, and dropout. Additionally, OnlineHD was configured with dimensional adjustment, bootstrap enabled, a learning rate of 0.035, and Gaussian distribution \(N(0,1)\). For ensemble models, \(N_L = 10\). The HDC model was evaluated with \(D_{total}\) values ranging from 10 to 10,000, while for the BoostHD model, the weak learner dimensionality (\(D_{wl}\)) was set to \(D_{total} / N_L\).  

    \begin{table*}[]
        \centering
        \caption{Accuracy (\%) Performance: Compares BoostHD`s accuracy with various baselines.}
            \label{tab: performance}
        \begin{tabular}{cccccccc}
        \hline
            Dataset & Adaboost & RF     & XGBoost & SVM    & DNN       & OnlineHD        & BoostHD         \\ \hline
            WESAD & 93.13 $\pm$ 0.01   & 97.13 $\pm$ 0.06  & 96.88 $\pm$ 0.01   & 93.12 $\pm$ 0.01  & 96.75 $\pm$ 1.05     & 96.37 $\pm$ 0.40           & \textbf{98.37} $\pm$ 0.32  \\
            Nurse Stress Dataset & 55.28 $\pm$0.01   & 59.35 $\pm$0.78  & 61.01 $\pm$0.01   & 60.99 $\pm$0.01  & 60.04 $\pm$0.06     & 61.37 $\pm$0.10           & \textbf{61.52} $\pm$0.07  \\
            Stress-Predict Dataset & 67.54 $\pm$ 0.01   & 67.76 $\pm$ 0.12  & 65.76 $\pm$ 0.01   & 67.30 $\pm$ 0.01  & 67.29 $\pm$ 0.07     & 65.79 $\pm$ 0.17          & \textbf{68.10} $\pm$ 0.09  \\
            \hline
        \end{tabular}
    \end{table*}

    \begin{table*}[]
        \centering
        \caption{Inference efficiency: Compares BoostHD`s Inference time ($10^{-5}$ seconds) with various baselines}
            \label{tab: inference}
        \begin{tabular}{cccccccc}
        \hline
                Dataset & Adaboost & RF       & XGBoost  & SVM      & DNN      & OnlineHD & BoostHD  \\ \hline
                WESAD & 46.3 & 38.5 & 47.6 & 108.3 & 37.0 & \textbf{7.57} & 11.0 \\
                Nurse Stress Dataset & 145.2 & 179.7 & 131.1 & 188.4 & 38.6 & 14.5 & \textbf{12.1} \\
                Stress-Predict Dataset & 63.4 & 91.5 & 58.7 & 265 & 43.7 & 13.2 & \textbf{12.0} \\ \hline
        \end{tabular}
    \end{table*}

    \subsection{Performance}
        In our comprehensive experimentation, we subjected the BoostHD algorithm to rigorous evaluation across three distinct datasets: WESAD\cite{wesad}, Nurse Stress Dataset\cite{nursestress}, and Stress-Predict Dataset\cite{Stress-PredictDataset}. The focal points of our assessment encompassed key metrics such as accuracy, inference time, and training time. Our findings unveiled the remarkable prowess of BoostHD, consistently attaining the highest accuracy levels across all three datasets, as presented in Table \ref{tab: performance}. Notably, when compared to OnlineHD on the WESAD\cite{wesad} dataset, the accuracy achieved by BoostHD remained outside the range of two standard deviations. This distinction was further underscored by OnlineHD's inability to reach a 98\% accuracy threshold in any of the ten independent trials, while BoostHD consistently surpassed this benchmark.
        
        BoostHD strategically combines in-memory learning, exemplified by HDC, with adaptive learning mechanisms within an ensemble framework, resulting in a significantly accelerated learning process, even when training is serialized. Empirical evaluations in a GPU environment demonstrated BoostHD's clear superiority over DNNs, achieving substantially faster processing speeds across the three benchmark datasets. In a CPU environment, BoostHD achieved remarkable computational efficiency, processing data 26 times faster than DNN counterparts. Additionally, optimization efforts tailored BoostHD for parallel processing, greatly improving inference efficiency Table \ref{tab: inference}. These enhancements were particularly effective when processing the relatively large input vectors in the Nurse Stress Dataset~\cite{nursestress} and Stress-Predict Dataset~\cite{Stress-PredictDataset}, resulting in a significant reduction in inference time. This firmly establishes BoostHD as the most time-efficient model across all evaluated datasets, particularly for the Nurse Stress Dataset~\cite{nursestress} and Stress-Predict Dataset~\cite{Stress-PredictDataset}.

    \subsection{Stability}
\begin{figure}[h]
        \centering
        \includegraphics[width=0.5\textwidth]{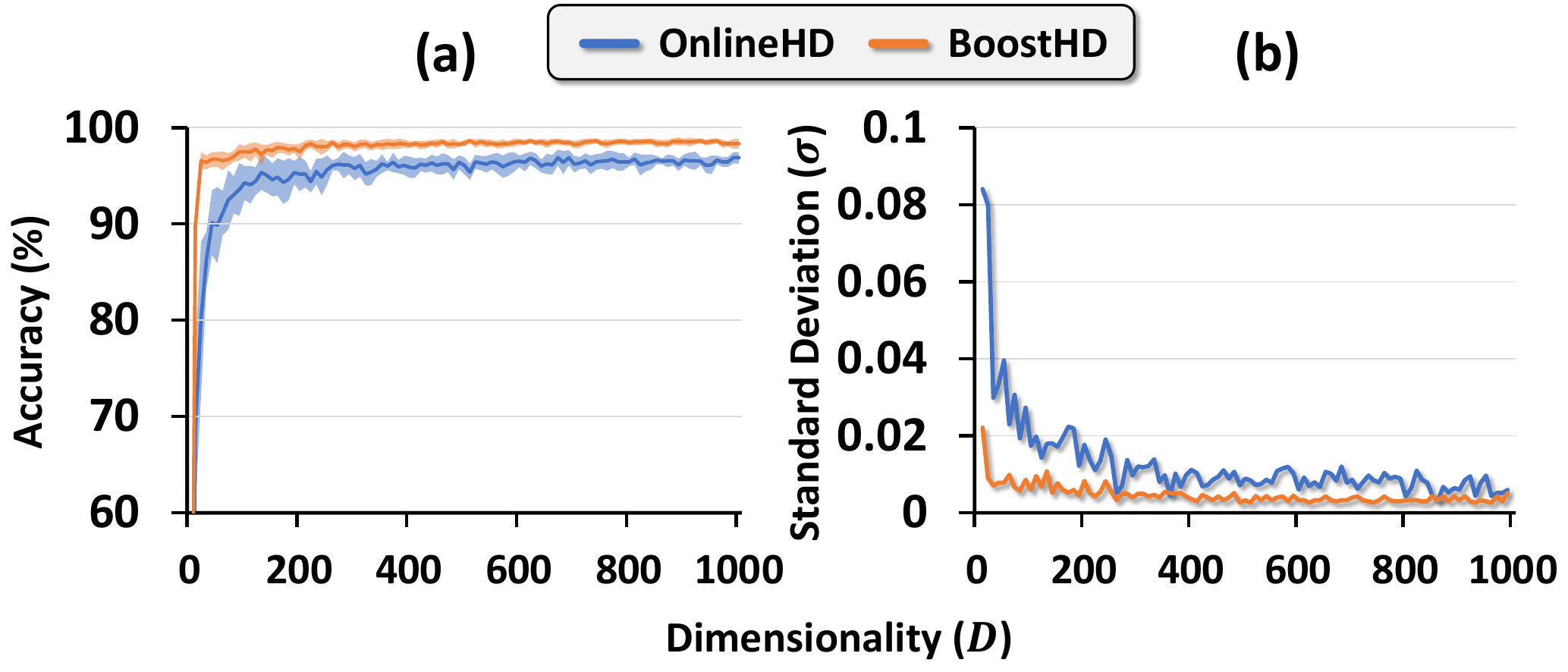}
        \caption{Analysis of the impact of $D$ on the stability of BoostHD and OnlineHD. (a), Accuracy of BoostHD and OnlineHD as a function of $D$, with error bars representing $\sigma$. (b), Variation of the $\sigma$ in (a) with respect to $D$.}
        \label{fig:stability}
    \end{figure}

        BoostHD demonstrates strong and consistent performance and exhibits stable convergence even in constrained environments. As illustrated in Figure \ref{fig:heatmap}, the accuracy of the weak learner experiences a steep decline when the minimum dimensionality requirement is unmet. By ensuring this minimum dimensionality and progressively increasing the value of $D$, we observe a clear separation between the two models within a given one standard deviation ($\sigma$), as shown in Figure \ref{fig:stability}(a). Specifically, the value of $\mu_\sigma$ for BoostHD stands at 0.0046, while for OnlineHD, it registers at 0.0127, signifying a roughly threefold disparity. This lower $\sigma$ value for BoostHD underscores its superior stability. In the context of BoostHD, it's notable that $\sigma$ scales proportionally with $1/N_L$ and inversely with $D$ provided that the baseline condition is maintained (as depicted in Figure \ref{fig:heatmap}(a), upper right). Moreover, when the value of $N_L$ surpasses a specific threshold ($k$), such as 50 as illustrated in Figure \ref{fig:heatmap}(a) (i.e., $N_L>k$), the $\sigma$ value remains consistently preserved.

    \begin{table*}[]
        \centering
        \caption{Person specific performance in accuracy (\%)}
        \label{tab: person}
        \begin{tabular}{cccccccc}
             \toprule
             & Left hands & Female & Age $\leq$ 25 & Age $\geq$ 30 & Height $\leq$ 170 & Height $\geq$ 185 & AVERAGE \\
             \hline
            Adaboost & 95.63 & 96.86 & 96.05 & 89.24 & 90.74 & 91.27 & 93.30 \\
            RF & 97.15 & 97.7 & 98.03 & 88.29 & 95.06 & 94.92 & 95.19 \\
            XGBoost & 98.73 & 98.12 & \textbf{98.68} & 91.41 & 95.06 & 94.24 & 96.04 \\
            SVM & 97.47 & 95.4 & 94.74 & 92.41 & 90.12 & \textbf{95.24} & 94.23 \\
            DNN & 98.42 & 95.4 & 94.41 & 88.92 & 92.59 & 93.57 & 93.89 \\
            OnlineHD & 98.73 & 97.8 & 94.08 & 92.41 & 92.59 & 94.6 & 95.04 \\
            BoostHD & \textbf{99.05} & \textbf{98.33} & 96.38 & \textbf{93.35} & \textbf{96.3} & 93.73 & \textbf{96.19} \\
            \hline
        \end{tabular}
    \end{table*}

    \subsection{Overfitting}
\begin{figure}[h]
        \centering
        \includegraphics[width=0.5\textwidth]{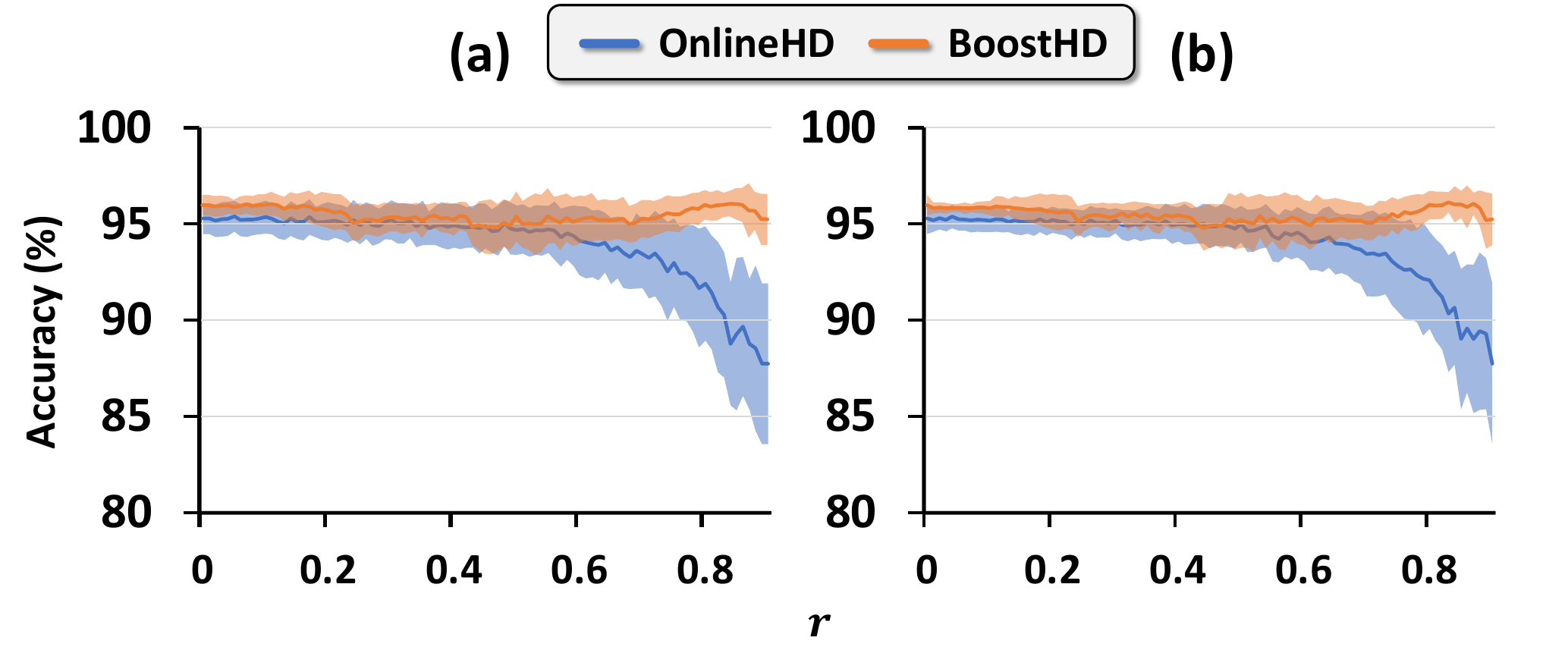}
        \caption{Analysis of the impact of r on overfitting BoostHD and OnlineHD. (a), (b) are the error variance in the overfitting with $D_{total}=1000$ and $ N_L = 10$. (b), $D_{total}=4000$ and $N_L = 10$.}
        \label{fig:overfit}
    \end{figure}

        BoostHD presents a noteworthy advantage over traditional HDC methods in its resilience to overfitting. To empirically assess this characteristic, we intentionally induce overfitting by crafting an imbalanced dataset ($\mathbf{D}$) through the generation of data subsets. The extent of overfitting, denoted as $r$, is quantified by creating subset data for all classes except the target class ($C_{target}$). This process yields the final dataset $\mathbf{D}$ as described in Equation \ref{eq:overfit_dataset}. We employ Macro accuracy as a more suitable metric for imbalanced datasets to ensure a fair performance evaluation that the varying sample counts per class do not skew. As the value of $r$ increases, OnlineHD experiences a noticeable decline in performance. In stark contrast, BoostHD consistently maintains its performance levels, demonstrating a robust ability to preserve stability, as depicted in Figure \ref{fig:overfit}.

    \begin{equation}
        \label{eq:overfit_dataset}
        \centering
        \mathbf{D}= \left\{
            \begin{aligned}
                \text{x}, & \; &\text{if } y=C_{\text{target}} \\ 
                \text{x} \times r, & \; & \text{if } y \neq C_{\text{target}}
            \end{aligned}
        \right.
    \end{equation}
    
    \subsection{Robustness}
\begin{figure}[h]
        \centering
        \includegraphics[width=0.5\textwidth]{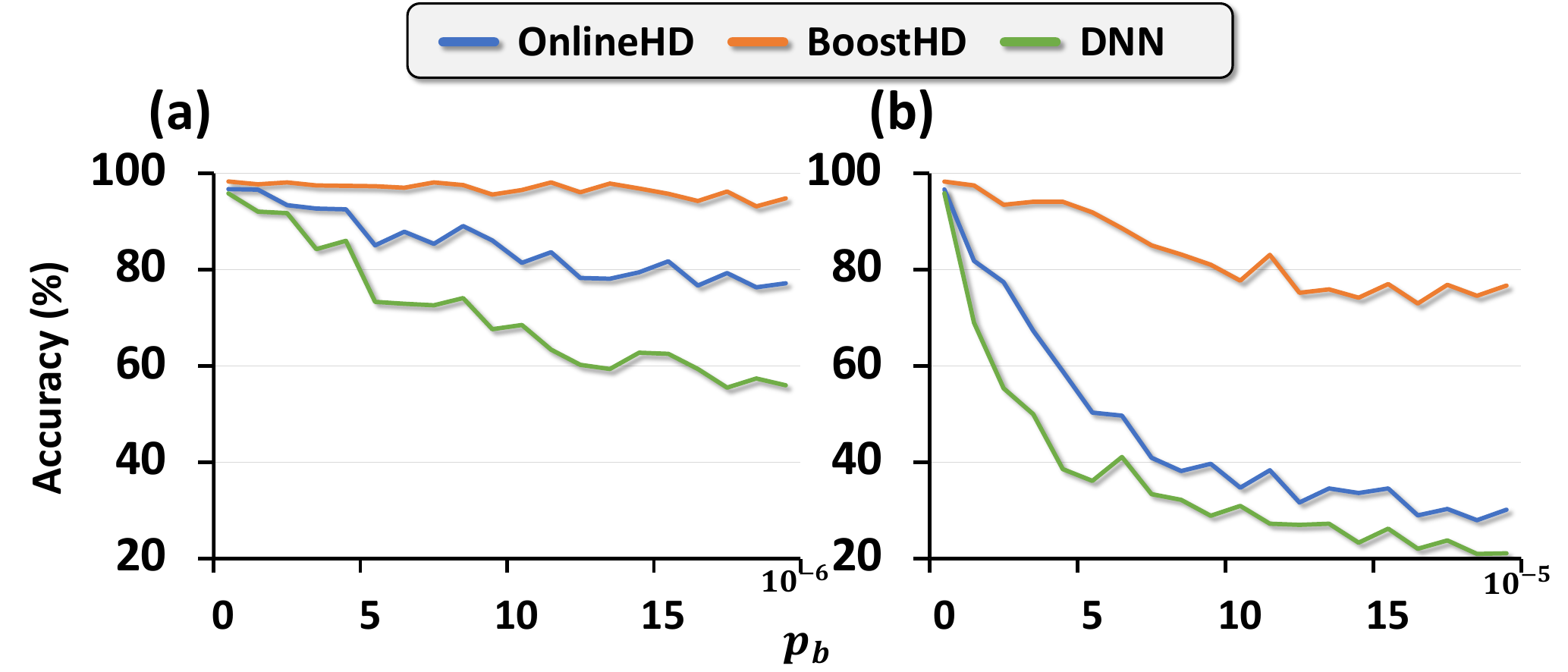}
        \caption{Robustness Analysis of BoostHD at different levels of $p_b$. At point (a), BoostHD experiences a mere 3\% decrement in performance, which is 6 and 13 times less pronounced than OnlineHD and DNN. Furthermore, at point (b), BoostHD manifests a 21\% reduction in performance, a figure that is 3 and 3.5 times less than the corresponding performance drops in OnlineHD and DNN.}
        \label{fig:robust}
    \end{figure}

        We assess the robustness of models against bitflip noise to explore in wearable device originating from hardware components according to its probability of bit flip error $p_b$. In Figure~\ref{fig:robust}, our experiments spanned two distinct ranges of $p_b$, given the substantial impact of this parameter on performance. Our experimental scope was deliberately limited to a narrow range of $p_b$. Despite this inherent constraint and our meticulous approach of conducting 100 independent trials to ensure statistical validity, we observed occasional peaks in the graph. Nevertheless, a discernible trend as a function of $p_b$ persisted. Within the range where $p_b = 10^{-5}$, as illustrated in Figure~\ref{fig:robust}(a), BoostHD incurred an accuracy loss of no more than 5.7\%. This represents approximately one-fourth of the loss observed in OnlineHD and roughly one-seventh of that observed in DNN. To statistically validate the accuracies concerning varying $p_b$, we employed the Median Absolute Deviation (MAD) as a measure of robustness (defined as $\text{MAD} = \text{median}(|X_i - \text{median}(X)|)$). In Figure~\ref{fig:robust}(a), the MAD for BoostHD was quantified at 0.024, which is six times lower than that of OnlineHD (0.1454) and four times lower than that of DNN (0.083). In Figure~\ref{fig:robust}(b), BoostHD exhibited a MAD of 0.005, which is three times lower than OnlineHD (0.015) and a substantial 30 times lower than DNN (0.1509). These results are compelling evidence of BoostHD's superior robustness compared to the other methods. 
            
    \subsection{Bias and reliability for person-specific groups}
        
        We carefully segmented subjects based on various attributes in WESAD\cite{wesad}, including hand preference, gender, age, and height. This stratification resulted in subsets with specific subject characteristics, such as left-hand preference, female gender, age groups, and height categories. Our primary objective was to assess how well models performed across individuals with diverse characteristics, a crucial consideration for healthcare applications to ensure fairness and accuracy. Notably, our BoostHD model consistently outperformed other models in all categories \ref{tab: person}, except for two cases where it ranked second. This underscores the potential of combining boosting methods and HDC, especially in healthcare, where equitable performance is essential.

\section{Conclusion}
We introduced BoostHD, a novel method that integrates hyperdimensional computing with boosting to create a robust ensemble model. BoostHD effectively overcomes limitations of traditional HDC by efficiently utilizing high-dimensional spaces and preventing overfitting. Our experiments on healthcare datasets demonstrated that BoostHD consistently outperforms existing methods in accuracy, efficiency, stability, and robustness. These results confirm BoostHD's potential in critical domains where reliability and precision are essential.

\section*{Acknowledgements}
This work was supported in part by the DARPA Young Faculty Award, the National Science Foundation (NSF) under Grants \#2127780, \#2319198, \#2321840, \#2312517, and \#2235472, the Semiconductor Research Corporation (SRC), the Office of Naval Research through the Young Investigator Program Award, and Grants \#N00014-21-1-2225 and \#N00014-22-1-2067, Army Research Office Grant \#W911NF2410360. Additionally, support was provided by the Air Force Office of Scientific Research under Award \#FA9550-22-1-0253, along with generous gifts from Xilinx and Cisco.

{
\bibliographystyle{IEEEtran}
\bibliography{mybibliography}
}

\end{document}